\documentclass[times, review, 10pt]{elsarticle}

\usepackage{amssymb}
\usepackage{amsmath}
\usepackage{multirow}
\usepackage{graphicx}
\usepackage{setspace}
\usepackage{float}

\journal{Pattern Recognition}

\begin{document}

\begin{frontmatter}


\title{Overcoming Occlusions in the Wild: A Multi-Task Age Head Approach to Age Estimation\tnoteref{label1}}



\author[inst1]{Waqar Tanveer\corref{cor1}}
\ead{wtan@unileon.es}
\author[inst2]{Laura Fernández-Robles}
\ead{lferr@unileon.es}
\author[inst1]{Eduardo Fidalgo}
\ead{eduardo.fidalgo@unileon.es}
\author[inst1]{Víctor González-Castro}
\ead{victor.gonzalez@unileon.es}
\author[inst1]{Enrique Alegre}
\ead{enrique.alegre@unileon.es}


\affiliation[inst1]{Universidad de León={Electrical, Systems and Automation Engineering, Universidad de León,},
            addressline={Campus de Vegazana s/n},
            postcode={24071},
            state={León},
            country={Spain}}

\affiliation[inst2]{Universidad de León={Department of Mechanical, Computer Science and Aerospace Engineering, Universidad de León,},
            addressline={Campus de Vegazana s/n},
            postcode={24071},
            state={León},
            country={Spain}}            

\cortext[cor1]{Corresponding author}

\title{}



%

\begin{abstract} 
Facial age estimation has achieved considerable success under controlled conditions. However, in unconstrained real-world scenarios, which are often referred to as 'in the wild', age estimation remains challenging, especially when faces are partially occluded, which may obscure their visibility. To address this limitation, we propose a new approach integrating generative adversarial networks (GANs) and transformer architectures to enable robust age estimation from occluded faces. We employ an SN-Patch GAN to effectively remove occlusions, while an Attentive Residual Convolution Module (ARCM), paired with a Swin Transformer, enhances feature representation. Additionally, we introduce a Multi-Task Age Head (MTAH) that combines regression and distribution learning, further improving age estimation under occlusion. Experimental results on the FG-NET, UTKFace, and MORPH datasets demonstrate that our proposed approach surpasses existing state-of-the-art techniques for occluded facial age estimation by achieving an MAE of $3.00$, $4.54$, and $2.53$ years, respectively.  

\end{abstract}




\end{frontmatter}



\section{Introduction} 

Estimating age from facial images has become a prominent yet challenging research topic, driven by its relevance to practical applications such as visual surveillance \cite{li2024gaitage}, forensics \cite{koch2025deep}, human-computer interaction (HCI) \cite{ruiz2024children}, and social media platforms \cite{mirabet2024facial}. Age estimation involves several intrinsic and extrinsic factors, making it a complex problem. Intrinsic factors, such as genetics, gender, and ethnicity, influence individual aging patterns and introduce variability in facial appearance \cite{angulu2018age}. Furthermore, imbalances in existing age estimation datasets, with uneven sample sizes across age groups, genders, and ethnicities, complicate the development of fair and unbiased systems \cite{yang2021delving}. Extrinsic challenges, such as image quality, illumination, facial expression, pose variation, and occlusion, significantly influence the performance of facial age estimation \cite{angulu2018age}.  Despite recent advances in age estimation research, achieving higher accuracy in the wild environment remains a big challenge. 

Age estimation typically comprises two main components: a feature extractor and an age estimator. Early studies relied on handcrafted features, including geometry and texture features \cite{kwon1999age}, Biologically Inspired Features (BIF) \cite{guo2009human}, and Scattering Transform (ST) \cite{chang2015learning}. The age estimator, in turn, has been approached as either a classification or a regression problem. Common classification models for age estimation include Support Vector Machine (SVM) \cite{guo2009human}, Multilayer Perceptron (MLP) \cite{lanitis2004comparing}, and k Nearest Neighbors (kNN) \cite{gunay2008automatic}, while regression methods such as Support Vector Regression (SVR) \cite{guo2009human} and quadratic regression \cite{guo2008image} were employed in prior approaches. 

Over the past few years, deep learning techniques have been employed to learn facial features directly, primarily through Convolutional Neural Networks (CNNs) \cite{chen2023daa,hiba2023hierarchical}. Several approaches used 2D CNNs to estimate age from facial images by combining them with domain adaptation \cite{singh2020deep} and attention mechanisms \cite{wang2022improving} to enhance the robustness and generalization of age estimation models. Recently, transformer-based approaches have gained significant attention from the research community for their potential in computer vision applications after their success in natural language processing \cite{vaswani2017attention}. Transformers have shown tremendous success over CNNs in image classification, 3D object detection, and image segmentation \cite{khan2022transformers}. Some approaches have also explored transformers with multi-task learning for face recognition, expression recognition, age estimation, and gender recognition \cite{qin2023swinface,kuprashevich2023mivolo}.  

Although prior studies have improved performance under controlled conditions \cite{shou2025masked,hiba2023hierarchical}
, accuracy in unconstrained scenarios (i.e., “in the wild” and "occluded")  remains unsatisfactory \cite{thorley2022estimates,barcic2022age}. 
Facial images captured in real-world environments are often affected by various factors such as lighting conditions and occlusions, which degrade the quality and clarity of the images \cite{elkarazle2022facial}. In particular, the widespread use of face masks following the COVID-19 pandemic has introduced substantial occlusions in facial images \cite{wong2022face}. Similarly, the facial images of individuals or crime suspects captured by CCTV cameras often exhibit considerable occlusions. In cases involving child sexual abuse material, masks are sometimes applied to the eyes to obscure identities \cite{chaves2023data}. These occlusions reduce the accuracy of age estimation due to the poor visibility of the faces \cite{jeuland2022assessment}. 
Only a limited number of studies in the current literature have considered occlusion while addressing the problem of age estimation \cite{barcic2022age,georgescu2022teacher,rothe2018deep,duan2017ensemble,chaves2023data,nam2023lca}. However, the results are not promising in comparison to those obtained in the controlled environment with full face visibility.


In this work, we introduce a framework aimed at mitigating the impact of occlusions in facial age estimation and improving its performance in the wild, followed by an extensive evaluation comparing our method to state-of-the-art approaches in occluded age estimation. Our main contributions in this paper are as follows:

\begin{itemize}

\item We propose a comprehensive deep learning-based framework for occluded facial age estimation to address the challenges posed by human age estimation in the wild.

\item We propose an Attentive Residual Convolution Module (ARCM) to enhance the feature representation for the age estimation task. Its purpose is to extract rich and meaningful information from the features of the Swin Transformer.

\item We design a Multi-Task Age Head (MTAH) that combines regression with distribution learning, thereby improving estimation accuracy even in heavily occluded images.




\end{itemize}

%

\section{Related Work}

\subsection{General Age Estimation}

In the early stages of age estimation, this problem was mostly addressed by using hand-crafted features. 
For instance, \citet{lanitis2002toward} employed the Active Appearance Model (AAM) to extract shape and appearance, while \citet{li2012learning} introduced ordinal discriminative features for facial age estimation. Similarly, \citet{yang2007demographic} combined the AdaBoost algorithm with Local Binary Patterns (LBP) for gender, ethnicity, and age classification. Later, \citet{guo2009human} proposed the Biologically Inspired Features (BIF), showing their effectiveness across multiple datasets. Previous studies have employed various handcrafted features, which required expert prior knowledge. However, no methods exist to assess the validity of such prior knowledge \cite{agbo2021deep}.

In recent years, deep learning approaches based on Convolutional Neural Networks (CNNs) have been widely used for the age estimation task. These CNNs automatically learn features from images, eliminating the need for handcrafted features. \citet{lin2022fp} developed a face-parsing network coupled with an attention mechanism to derive age from semantic features. 
A novel attention-based approach for facial age estimation was presented by \citet{hiba2023hierarchical}, involving multiple input image augmentations, aggregation of CNN embeddings through a Transformer-Encoder module, and age prediction using a regressor. Additionally, \citet{chen2023daa} utilized a Delta Age AdaIN (DAA) operation with a binary code transformer to capture inter-age feature differences, deriving a style map from mean and standard deviation values and applying age transfer learning for the final estimation.

\subsection{Occlusion-based Facial Age Estimation}

Although deep learning has significantly boosted age estimation accuracy, practical deployment still faces challenges under real-world conditions, particularly due to occlusions. Only a few studies have tried to address this problem. \citet{chaves2023data} improved age estimation for children and young adults by training model on a combination of non-occluded and partially occluded images, employing a compact Soft Stage-wise Regression Network (SSR-Net) for hardware efficiency. \citet{yadav2014recognizing} investigated the facial cues used by humans to estimate age, ethnicity, and gender across various age groups, along with the impact of different facial features, such as the eyes and mouth, on age estimation. In another work, \citet{ye2018privacy} proposed a privacy-preserving framework targeting images with masked eyes, using a VGG-16 backbone and an attention module for discriminative feature learning.  

\citet{barcic2022age} created occluded images by adding artificial masks to the face and utilized the EfficientNet-B3 CNN model for facial age estimation.\citet{geng2016label} implemented Label Distribution Learning (LDL) to model the relation between facial images and age labels, incorporating a cost-sensitive loss to enhance the performance of age classifier on imbalanced face datasets. \citet{nam2023lca} introduced a Low-Complexity-Attention (LCA) GAN network for age estimation of people wearing face masks, employing Conditional GAN (CGAN) to remove mask occlusions from the nose and mouth regions and incorporating content and edge loss to reconstruct hidden facial features. A multi-task learning framework for age, gender, and facial expression recognition using knowledge distillation with a teach-student training strategy was proposed to encounter the effects of occlusion \cite{georgescu2022teacher}. The teacher model was trained on fully visible faces, whereas the student model was trained on half-occluded faces using triplet loss.

Despite these efforts, a substantial performance gap persists between non-occluded and occluded faces, resulting in suboptimal accuracy in real-world scenarios (in the wild). To mitigate this issue, we propose a two-phase framework that leverages a combination of GANs and transformers to enhance occluded facial age estimation, making our approach effective in controlled and uncontrolled environments.


\section{Methodology}

\begin{figure}[htbp]
\centerline{\includegraphics[width=0.95\linewidth]{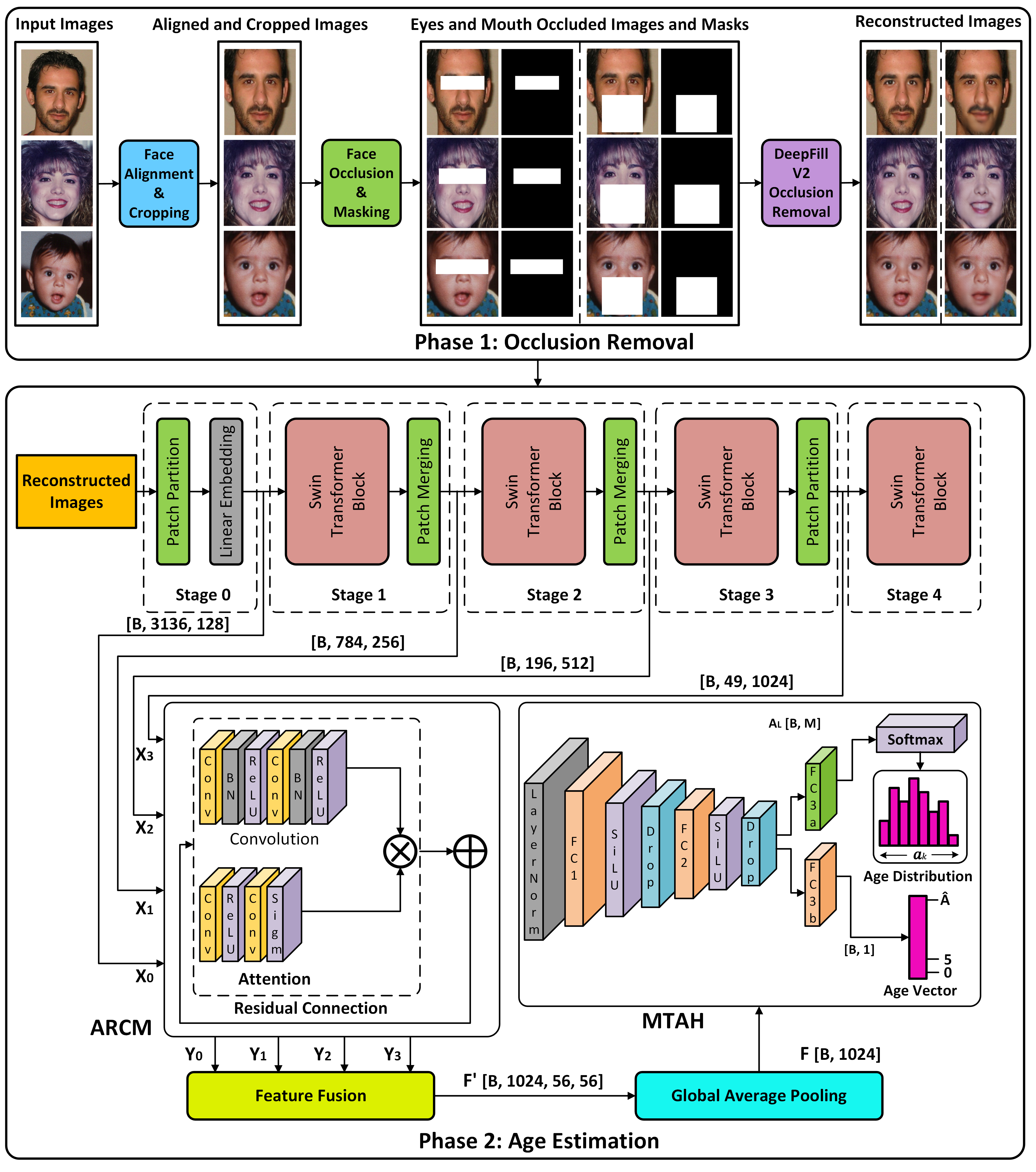}} 
\caption{An overview of our two-phase proposed approach for occluded facial age estimation. In phase 1, the input images are aligned and cropped using dlib 68-point facial landmark extraction. DeepFillv2 is utilized to remove occlusion from the occluded images. In the second phase, the reconstructed images are given as input to the Swin Transformer for feature extraction. The features from the Swin Transformer are further processed by an Attentive Residual Convolution Module (ARCM), which refines feature details using attention and residual learning. The refined features are fused and input to a Multi-Task Age Head (MTAH) for age estimation.}
\label{fig:complete_framework}
\end{figure}

To overcome the effect of occlusion in facial images for age estimation, we propose a two-phase framework. Our complete framework is shown in Fig. \ref{fig:complete_framework}. The first phase removes the occlusions from the occluded images, whereas the second phase performs the age estimation. The details of the methodology are given as follows.

\subsection{Occlusion Removal}

\subsubsection{Face Alignment and Cropping}
\label{sec:Face_Alignment_Cropping}

The input images are first aligned and cropped using facial landmark extraction implemented using the dlib toolkit \cite{king2009dlib}. The central point between the eyes is calculated by taking the average of the coordinates of the left and right eyes. After that, the angle of rotation is computed from the eye coordinates using Eq. \eqref{eq:rotation_angle}, which is then used to generate the rotation matrix centered at the midpoint between the eyes. Finally, the images are aligned using an affine transformation. Once the images are aligned, the face is cropped using the frontal face detection feature of dlib.

\begin{equation}
\theta = \tan^{-1}\left(\frac{\Delta y}{\Delta x}\right) \times \frac{180}{\pi}
\label{eq:rotation_angle}
\end{equation}
where
\[
\Delta y = y_{\text{right\_eye}} - y_{\text{left\_eye}}, \quad \Delta x = x_{\text{right\_eye}} - x_{\text{left\_eye}},
\]
representing the differences between vertical and horizontal image coordinates, respectively, for both left and right eyes.

\subsubsection{Face Occlusion and Masking}
\label{sec:Face_Occlusion_Masking}
Since there are no datasets available for age estimation that contain occluded facial images, we created occluded variants from existing datasets. Occlusions were synthetically applied based on the positions of facial landmarks detected using the 68-point landmark of dlib. Two types of rectangular occlusions were designed: the first one covers the eyes and surrounding areas, simulating sunglasses or eye masks, whereas the second one covers the mouth area, resembling face masks. To occlude the eyes, the coordinates of the left eye (landmark 36) and right eye (landmark 45) are extracted, capturing their x and y positions. A rectangle is calculated using these coordinates, with boundaries defined by the minimum and maximum x and y values of the eye points, extended by 25 pixels horizontally and 20 pixels vertically to fully cover the eye area. This rectangle, centered around the eyes, is filled with solid white (255, 255, 255) pixels, effectively occluding the eye region. Additionally, corresponding binary masks are generated by keeping the eye occluded region with (255, 255, 255) pixels and filling the remaining region with black (0,0,0) pixels. 

\begin{figure}
\centerline{\includegraphics[width=0.95\linewidth]{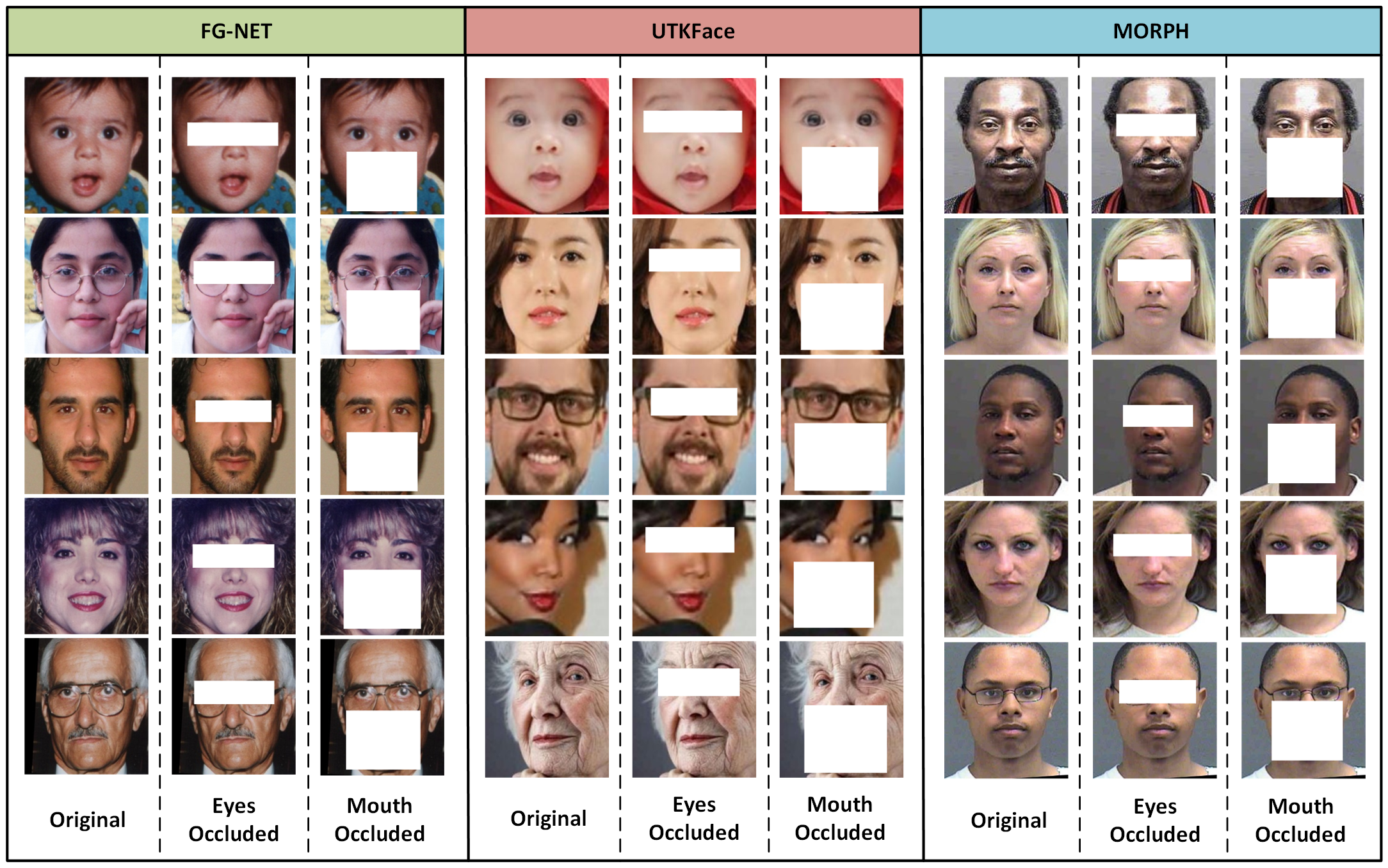}} 
\caption{Image samples and their corresponding eye-occluded and mouth-occluded versions. Columns 1, 2, and 3 show examples from the FG-NET, UTKFace, and MORPH datasets, respectively.}
\label{fig:Examples_occlusions}
\end{figure}

A similar procedure is carried out for the mouth occlusion by extracting the coordinates of the left mouth corner (landmark 48) and right mouth corner (landmark 54). A rectangle is calculated around the mouth area, with boundaries defined by the minimum and maximum x and y values of these mouth points, extended by 40 pixels horizontally and 55 pixels vertically for the FG-NET and MORPH datasets and by 45 and 60 pixels for the UTK dataset to fully cover the mouth region. The binary mask for the mouth region is generated in the same way as the mask for the eyes.
Fig. \ref{fig:Examples_occlusions} shows the original, eye-occluded, and mouth-occluded images.

\subsubsection{Occlusion Removal}
\label{sec:Occlusion_Removal_Methodology}
To remove the occlusion from the facial images and restore their original appearance, we utilized DeepFillv2 \cite{yu2019free}, a state-of-the-art image inpainting method. The architecture of DeepFillv2 is divided into two stages. In the first stage, the set of occluded images $I_{oc} = \lbrace{I_{oc1}, I_{oc2},..., I_{ocn}}\rbrace$ and their corresponding binary masks $M = \lbrace{M_1, M_2,..., M_n}\rbrace$ are fed into the Coarse Network, which uses gated convolutions to extract low-level features and generate an initial coarse reconstruction of the occluded facial regions as given in Eq. \eqref{eq:feature_map} and Eq. \eqref{eq:coarse_output}.

\begin{equation}
F_{\text{out}} = \phi(W_f * [I_{oc}, M]) \odot S(W_g * [I_{oc}, M])
\label{eq:feature_map}
\end{equation}
where  $F_{out}$ denotes the output feature map, \( W_f \) is the feature filter, \( W_g \) is the gating filter, \( [I_{oc}, M] \) is the input feature map the channel-wise concatenation of the input image and its mask, \( * \) denotes convolution, \( \odot \) represents element-wise multiplication, \( S \) is the sigmoid activation function, and \( \phi \) is the ReLU activation function used in this work.

\begin{equation}
I_{\text{coarse}} = G_{\text{coarse}}(I_{oc}, M) = W_{\text{out}} * F_{\text{out}}
\label{eq:coarse_output}
\end{equation}
where \(G_{\text{coarse}}\) represents the coarse generator function, \( W_{\text{out}} \) is a 1\(\times\)1 convolution that transforms \( F_{\text{out}} \) to the 3-channel coarse reconstruction \( I_{\text{coarse}} \).

The coarse output \( I_{\text{coarse}} \) is then passed to the second stage, which consists of a two-branch Refinement Network and a Contextual Attention Module. In the Refinement Network, gated convolutions dynamically adapt to the binary mask, prioritizing non-occluded facial features (e.g., eyes, nose, mouth). The Contextual Attention Module reconstructs occluded regions by transferring features from non-occluded patches, as in Eq. \eqref{eq:contextual_attention}.

\begin{equation}
p_{\text{out}} = \sum_j a_j p_j = \sum_j \left[ \frac{\exp(\langle \text{norm}(p_{\text{missing}}), \text{norm}(p_j) \rangle)}{\sum_k \exp(\langle \text{norm}(p_{\text{missing}}), \text{norm}(p_k) \rangle)} \right] p_j
\label{eq:contextual_attention}
\end{equation}
where \( p_{\text{out}} \) is the reconstructed patch for an occluded region (\( M = 0 \)), \( p_{\text{missing}} \) is the original patch from the occluded area in the feature map \( F \), \( p_j \) and \( p_k \) are patches from non-occluded regions (\( M = 1 \)), \( \text{norm}(p) = \frac{p}{\| p \|_2} \), and \( \langle \cdot, \cdot \rangle \) computes cosine similarity. The attention weights \( p_j \) prioritize patches with similar facial features. These patches \( p_{\text{out}} \) replace occluded regions in \( F \) to form the feature map \( F'_{\text{out}} \), which is further used  to produce the 3-channel refined output \( I_{\text{recon}} \) as given in Eq. \eqref{eq:reconstructed}.

\begin{equation}
I_{\text{recon}} = G_{\text{refine}}(I_{\text{coarse}}, M) = H(F'_{\text{out}}, F_{\text{gated}})
\label{eq:reconstructed}
\end{equation}
where \( G_{\text{refine}} \) is the refinement network, \( F'_{\text{out}} \) is the feature map with \( p_{\text{out}} \) patches integrated, \( F_{\text{gated}} \) is the feature map from the gated convolution branch processing \( [I_{\text{coarse}}, M] \), and \( H \) represents subsequent layers combining \( F'_{\text{out}} \) and \( F_{\text{gated}} \).

The final output of the Refinement Network is a reconstructed image in which the occlusions are eliminated, restoring the original facial appearance. The Spectral-Normalized (SN-PatchGAN) Discriminator is used to discriminate between the input image and the reconstructed image by using the loss functions defined in Eq. \eqref{eq:hinge_loss_g} and Eq. \eqref{eq:hinge_loss_d}.

\begin{equation}
L_G = -\mathbb{E}_{I_{oc} \sim P_{I_{oc}}(I_{oc})} [D_{\text{SN}}(G(I_{oc}))]
\label{eq:hinge_loss_g}
\end{equation}

\begin{equation}
L_{D_{\text{SN}}} = \mathbb{E}_{I \sim P_{\text{I}}(I)} [\phi(1 - D_{\text{SN}}(I))] + \mathbb{E}_{I_{oc} \sim P_{I_{oc}}(I_{oc})} [\phi(1 + D_{\text{SN}}(G(I_{oc})))]
\label{eq:hinge_loss_d}
\end{equation}
where \( D_{\text{SN}} \) represents the spectral-normalized discriminator, \( G \) is the image generator that generates the reconstructed image \(I_{\text{recon}}\), \( P_{\text{I}}(I) \) is the distribution of real images \( I \), \( P_{I_{oc}}(I_{oc}) \) is the distribution of occluded images \(I_{oc}\), and \(\mathbb{E}\) denotes the expectation, computing the average over the specified distribution. \( D_{\text{SN}} \) tries to distinguish real patches from generated ones, while \( L_G \) encourages \( G \) to produce better reconstructed images.

\subsection{Age Estimation}
\label{sec:Age_Estimation_Methodology}

In Fig. \ref{fig:complete_framework}, phase 2 shows our framework for age estimation from facial images. Our age estimation framework mainly consists of three main blocks, which include a Swin Transformer (STF), an Attentive Residual Convolution Module (ARCM), and a Multi-Task Age Head (MTAH).

\subsubsection{Swin Transformer}
\label{sec:Swin_Transformer}

We use the Swin Transformer (STF) \cite{liu2021swin} as the backbone network to extract features with its classification head removed. Stages 0-3 are used for feature extraction to capture fine-grained details, such as skin texture and wrinkles, critical for age estimation. The input images \( I_n \in \mathbb{R}^{B \times C \times H \times W} \) (where \( n \) represents the total number of images, \( B \) is the batch size, \( C = 3 \) is the number of channels, and \( H = 224 \), \( W = 224 \) are the height and width) are processed through a patch embedding layer. This layer applies a convolutional operation (\( 4 \times 4 \), stride 4) to divide each image \( I \) into non-overlapping \( 4 \times 4 \) patches (resulting in \( 56 \times 56 = 3136 \) patches) and projects them into a hidden dimension of 128 channels, yielding \( X_0 \in \mathbb{R}^{B \times 3136 \times 128} \). Stage 1 applies 2 STF blocks with windowed attention (3 heads) followed by patch merging (reducing to \( 28 \times 28 = 784 \) patches), producing \( X_1 \in \mathbb{R}^{B \times 784 \times 256} \). Stage 2 uses patch merging (to \( 14 \times 14 = 196 \) patches) and 2 STF blocks (6 heads) to yield \( X_2 \in \mathbb{R}^{B \times 196 \times 512} \). Stage 3 applies patch merging (to \( 7 \times 7 = 49 \) patches) and 18 STF blocks (12 heads) for \( X_3 \in \mathbb{R}^{B \times 49 \times 1024} \). The selected features from Stage 0 to Stage 3 are reshaped into spatial maps (\( \mathbb{R}^{B \times C_i \times \sqrt{N_i} \times \sqrt{N_i}} \), where \( C_i = 128, 256, 512, 1024 \) and \( N_i = 3136, 784, 196, 49 \)), producing resolutions of \( 56 \times 56 \), \( 28 \times 28 \), \( 14 \times 14 \), and \( 7 \times 7 \) pixels, respectively.

\subsubsection{Attentive Residual Convolution Module}
\label{sec:ARCM}

To improve the quality of feature representation, we propose an Attentive Residual Convolution Module (ARCM), which combines multiple convolutional and attention layers with residual connections to refine the features extracted from the STF stages. 

The output features from the Swin Transformer (STF) stages \( X_0, X_1, X_2, X_3 \) are passed through the first convolutional block, applying a \( 3 \times 3 \) convolution with \( C \) input and output channels and padding of 1 to maintain spatial dimensions, followed by batch normalization and ReLU activation, producing an intermediate feature map in \( \mathbb{R}^{B \times C \times H_i \times W_i} \). This is followed by a second convolutional block with the same \( 3 \times 3 \) convolution, batch normalization, and ReLU activation, producing a refined feature map \( F_{\text{conv}} \in \mathbb{R}^{B \times C \times H_i \times W_i} \). The attention mechanism then processes \( F_{\text{conv}} \), applying a \( 1 \times 1 \) convolution to reduce channels to \( C/8 \) (16, 32, 64, 128 channels for \( C = 128, 256, 512, 1024 \), respectively), followed by ReLU activation, then a \( 1 \times 1 \) convolution to restore channels to \( C \), and a sigmoid activation to produce attention weights \( A \in \mathbb{R}^{B \times C \times H_i \times W_i} \) with values between 0 and 1. The final output \( Y \in \mathbb{R}^{B \times C \times H_i \times W_i} \) is obtained by element-wise multiplying \( F_{\text{conv}} \) with the attention weights \( A \) and adding the input feature map \( X_i \) as a residual connection, enhancing the feature representation as given in Eq. \eqref{eq:element_wise_multiplication}
\begin{equation}
Y = F_{\text{conv}} \odot A + X_i
\label{eq:element_wise_multiplication}
\end{equation}
This process is applied to Stage 0 (\( C = 128 \), \( H_i = W_i = 56 \)), Stage 1 (\( C = 256 \), \( H_i = W_i = 28 \)), Stage 2 (\( C = 512 \), \( H_i = W_i = 14 \)), and Stage 3 (\( C = 1024 \), \( H_i = W_i = 7 \)), producing \( Y_0 \in \mathbb{R}^{B \times 128 \times 56 \times 56} \), \( Y_1 \in \mathbb{R}^{B \times 256 \times 28 \times 28} \), \( Y_2 \in \mathbb{R}^{B \times 512 \times 14 \times 14} \), and \( Y_3 \in \mathbb{R}^{B \times 1024 \times 7 \times 7} \) as outputs of the ARCM.

\subsubsection{Feature Fusion}
\label{sec:FF}
Feature fusion enhances the feature representation by combining the multi-scale features obtained from the ARCM, ensuring that valuable raw details are not lost in the process. It integrates \( Y_0, Y_1, Y_2, Y_3 \) (from ARCM outputs corresponding to Stages (0-3), producing an intermediate fused feature tensor \( F' \in \mathbb{R}^{B \times 1024 \times 56 \times 56} \). 

Each \( Y_n \) (\( Y_0 \in \mathbb{R}^{B \times 128 \times 56 \times 56} \), \( Y_1 \in \mathbb{R}^{B \times 256 \times 28 \times 28} \), \( Y_2 \in \mathbb{R}^{B \times 512 \times 14 \times 14} \), \( Y_3 \in \mathbb{R}^{B \times 1024 \times 7 \times 7} \)) is first processed through a \( 1 \times 1 \) convolution to project the channel dimension to 1024, followed by bilinear interpolation to upsample the spatial dimensions of \( Y_1, Y_2, Y_3 \) to match the largest resolution of \( 56 \times 56 \) from \( Y_0 \). The outputs are then concatenated along the channel dimension, resulting in a feature map of \( \mathbb{R}^{B \times 4096 \times 56 \times 56} \), which is reduced to \( F' \) via a final \( 1 \times 1 \) convolution after batch normalization and ReLU activation.

An average pooling operation is then applied to take the average across the \( 56 \times 56 \) spatial dimensions, resulting in the final fused feature tensor \( F \in \mathbb{R}^{B \times 1024} \) defined in Eq. \eqref{eq:Fused_Feature_Vector}.

\begin{equation}
F(b, c) = \frac{1}{56 \times 56} \sum_{h=0}^{55} \sum_{w=0}^{55} F'(b, c, h, w)
\label{eq:Fused_Feature_Vector}
\end{equation}
where \( b \) is the batch index, \( c \) is channel index ranging from 0 to 1023, and \( h, w \) are the spatial indices.

\subsubsection{Multi-Task Age Head}
\label{sec:MTAH}
To perform age estimation, we do not rely only on regression. We propose a Multi-Task Age Head (MTAH) module to transform high-dimensional fused features into a probabilistic distribution over age bins for age estimation, alongside a direct regression output. It takes as input a feature vector \( F \in \mathbb{R}^{B \times 1024} \) and processes this input through a sequence of operations. It starts by applying the normalization layer across the 1024 channels to stabilize and standardize the feature scales, mitigating the impact of varying feature magnitudes. This is followed by a fully connected (linear) layer that transforms the 1024-dimensional features while maintaining their dimensionality, followed by a SiLU (Sigmoid Linear Unit) activation to introduce nonlinearity and enhance the gradient flow. A dropout layer is also added with a $50\%$ probability to prevent overfitting by randomly zeroing out features during training. Another linear layer is used to further reduce the dimensionality from 1024 to 512, again followed by a SiLU activation for additional nonlinearity and a dropout layer with a $40\%$ probability. The last linear layers project the 512-dimensional feature vectors into two outputs: a single regression value for direct age prediction and \( M \) age logits \( A_L \in \mathbb{R}^{B \times M} \) corresponding to age bins, ranging from \( N \) to \( N + M - 1 \), where \( N \) is the starting index depending on the dataset (e.g., 0 for FGNET, 16 for MORPH). These logits represent unnormalized log-probabilities for each age bin and can be converted to normalized probabilities \( P \in \mathbb{R}^{B \times M} \), which sum to 1 for each sample, via a softmax operation. To form a predicted age distribution, the logits are processed through a log-softmax operation to produce log-probabilities \( P_{log} \in \mathbb{R}^{B \times M} \), where \( P_{log}(x, k) = \log P(x, k) \)

To bring the ground-truth ages \( A \in \mathbb{R}^{B \times 1} \) into distribution form, a target distribution \( T \in \mathbb{R}^{B \times M} \) is constructed using a Gaussian function centered at the ground-truth ages using Eq.~\eqref{eq:Target_Distribution}. 

\begin{equation}
T(x, k) = \exp\left(-\frac{(a_k - A(x))^2}{2 \sigma^2}\right)
\label{eq:Target_Distribution}
\end{equation}
where \( x \) indexes the samples in the batch (from 1 to \( B \)), \( k \) indexes the age bins (from \( N \) to \( N+M-1 \)  where N is the starting index, depending on the dataset), \( a_k \) represents the age value corresponding to bin \( k \), and \( \sigma \) represents the fixed standard deviation to control the spread. The distribution is normalized via \( T(x, k) = T(x, k) / \sum_{k} T(x, k) \) to ensure it sums to 1 per sample. After generating the target distribution for ground-truth ages and obtaining the classification and regression outputs, we used a weighted combination of two losses as a loss function for our proposed approach as given in Eq.~\eqref{eq:Loss}.

\begin{equation}
L_{age} = \alpha L_{\delta} + \beta L_{KL}
\label{eq:Loss}
\end{equation}
where \( L_{\delta} \) and \( L_{KL} \) are the Huber Loss and the KL-Divergence Loss defined in Eq.~\eqref{eq:Huber_Loss} and Eq.~\eqref{eq:KL_Loss} respectively. \( \alpha \) and \( \beta \) are the hyperparameters to balance both losses.

\begin{equation}
L_{\delta}(\hat{A}, A) = 
\begin{cases} 
\frac{1}{2} (\hat{A} - A)^2, & \text{if } |\hat{A} - A| \leq \delta \\
\delta |\hat{A} - A| - \frac{1}{2} \delta^2, & \text{if } |\hat{A} - A| > \delta 
\end{cases}
\label{eq:Huber_Loss}
\end{equation}
where \( \hat{A} \) is the predicted age, \( A \) is the true age, and \( \delta \) is a positive threshold factor that determines the transition between the quadratic (L2) loss for small errors and the linear (L1) loss for larger errors.

\begin{equation}
L_{KL} = \sum_{k} T(x, k) \cdot (\log T(x, k) - \log P(x, k))
\label{eq:KL_Loss}
\end{equation}
where \( P(x, k) \) represents the predicted probability for sample \( x \) and age bin \( k \), obtained by applying a softmax operation to the classification logits \( A_L \in \mathbb{R}^{B \times M} \), and the summation is over the age bins as defined by the dataset.

\section{Experimentation} 
\label{sec:Experimentation}

In this Section, two types of experiments are performed. First, the occlusion removal pipeline from phase 1 of our proposed approach is implemented to remove occlusions from occluded eye and mouth images, yielding reconstructed eye and mouth images. Next, the pipeline for phase 2 is implemented to perform occluded facial age estimation on the reconstructed images. The details of the experiments are as follows:

\begin{figure}[]
\centerline{\includegraphics[height= 8 cm, width=11cm]{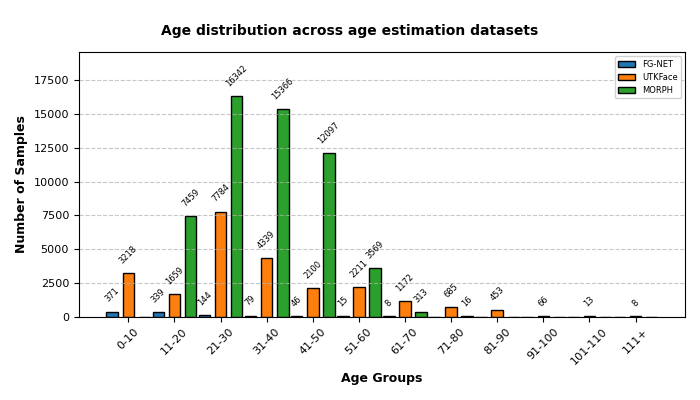}}
\caption{Distribution of image samples according to age groups in FG-NET, UTKFace, and MORPH datasets.}
\label{fig:Combined_Distribution}
\end{figure}


\subsection{Datasets} \label{sec:Datasets}

To evaluate our proposed approach, we have used three well-known public benchmark age estimation datasets, FG-NET \cite{panis2016overview}, UTKFace \cite{zhang2017age}, and MORPH \cite{ricanek2006morph}. The distribution of image samples across age groups with 10 or fewer samples per group for each dataset is shown in Fig. \ref{fig:Combined_Distribution}.

\textbf{FG-NET} dataset contains 1,002 facial images from 82 individuals, with ages spanning from 0 to 69 years. It presents challenges due to considerable differences in pose, facial expressions, and illumination across the images.

\textbf{UTKFace} dataset contains over 20,000 face images annotated with age (0–116 years), gender, and ethnicity (White, Black, Asian, Indian, Others). It features a diverse but imbalanced ethnic distribution, with approximately 50\% White and smaller proportions of other groups. The images exhibit varied poses, expressions, and lighting conditions, making them suitable for tasks like age and ethnicity prediction.

\textbf{MORPH} dataset comprises over 55,000 images from more than 13,000 individuals of diverse racial backgrounds. Each image is accompanied by the chronological age, with ages spanning 16 to 77 years and an average age of 33 years. Over 96\% of the facial images represent individuals of European or African descent, reflecting a highly imbalanced ethnic composition.

\subsection{Evaluation Metrics}
\label{sec:Evaluation_Metrics}

Three metrics are used to evaluate our proposed approach. Peak Signal-to-Noise Ratio (PSNR) and Structural Similarity Index (SSIM) are utilized to measure the quality of reconstructed images produced by DeepFillv2, while Mean Absolute Error (MAE) is used to assess the performance of age estimation.

\textbf{PSNR} is employed to evaluate the quality of a reconstructed image \( I_{rec}(u,v) \) by comparing it to the original image \( I(u,v) \). The equation for PSNR is given in Eq. \eqref{eq:PSNR}.
\begin{align}
PSNR = 10 \hspace{1mm} \log_{10} \left( \dfrac{P^2}{MSE} \right)
\label{eq:PSNR}
\end{align}
where \( P \) denotes the maximum possible pixel value in the image, \( (u,v) \) are the image coordinates,   and \( MSE \) is the mean squared error between the images.

\textbf{SSIM} is a perceptual metric utilized to assess the similarity between two images, \( I \) and \( I_{rec} \), by evaluating their luminance, contrast, and structural components. SSIM is defined as the product of these three components, which assess perceived changes in brightness, contrast, and structural information, respectively, as given in Eq. \eqref{eq:SSIM}. The luminance term compares the mean intensities, the contrast term compares the standard deviations, and the structure term evaluates the covariance between the images. The combined SSIM formula is:
\begin{align}
SSIM(I, I_{rec}) = \dfrac{(2 \mu_I \mu_{I_{rec}} + C_1)(2 \sigma_I \sigma_{I_{rec}} + C_2)}{(\mu_I^2 + \mu_{I_{rec}}^2 + C_1)(\sigma_I^2 + \sigma_{I_{rec}}^2 + C_2)}
\label{eq:SSIM}
\end{align}
where \( \mu_I \) and \( \mu_{I_{rec}} \) are the mean intensities of the images \( I \) and \( I_{rec} \), \( \sigma_I \) and \( \sigma_{I_{rec}} \) are their standard deviations, and \( \sigma_{II_{rec}} \) is the covariance between the images. The constants \( C_1 \) and \( C_2 \) are included in the denominators to prevent division by zero and in the numerators to stabilize the luminance and contrast comparisons, ensuring robust results even when mean intensities or standard deviations are small.

\textbf{MAE} computes the average of absolute differences between the actual age $A$ and the estimated age $\hat{A}$. The formula for the MAE is given in Eq. \eqref{eq:MAE}.
\begin{align}
MAE = \dfrac{1}{T}\sum_{i=1}^{T} \vert A - \hat{A} \vert
\label{eq:MAE}
\end{align}
where T is the total number of images.

\subsection{Implementation Details}
\label{sec:Implementation_Details}

First, we use the dlib face detector \cite{king2009dlib} to refine the age estimation datasets by excluding the images in which faces are not present. After cleaning the datasets, face cropping and alignment are performed using dlib's facial landmark detection to exclude the unnecessary area from the facial images. Each dataset is then randomly divided into training, validation, and test sets, containing 60\%, 20\% and 20\%, respectively, of the dataset samples. Table \ref{tab:dataset_metrics} shows the number of images used for training, validation, and testing for each dataset.

\subsubsection{Occlusion Removal} \label{sec:Occlusion_Removal_Experimentation}

As there are no occluded datasets available for facial age estimation, we created the occluded versions of FG-NET, UTKFace, and MORPH. The white masks were created following the procedure described in Sections \ref{sec:Face_Alignment_Cropping} and \ref{sec:Face_Occlusion_Masking}.  We utilize the SN-Patch GAN from DeepFillv2 \cite{yu2019free} to remove occlusions from the images and reconstruct the images similar to the original ones. The input images and their corresponding binary masks are first resized to \(256\times256\). After that, the SN-Patch GAN is trained for \( 1 \times 10^{6} \) iterations with a learning rate of \( 1 \times 10^{-4} \) using Adam optimizer for both generator and discriminator for a batch size of 16 images. SSIM and PSNR are used to evaluate the generated images to compare with the original images.

\begin{table}[]
\centering
\scriptsize
\caption{Statistics about the number of images used for age estimation datasets}
\label{tab:dataset_metrics}
\begin{tabular}{|l|c|c|c|}
\hline
\textbf{\begin{tabular}[c]{@{}l@{}}No. of Images / \\ Datasets\end{tabular}} & \textbf{FG-NET} & \textbf{UTKFace} & \textbf{MORPH} \\ \hline
Train                                                                        & 585             & 13881            & 33097          \\ \hline
Validation                                                                   & 195             & 4627             & 11032          \\ \hline
Test                                                                         & 195             & 4627             & 11033          \\ \hline
Total                                                                        & 975             & 23135            & 55132          \\ \hline
\end{tabular}
\end{table}

\subsubsection{Age Estimation}
\label{sec:Age_Estimation_Experimentation}

For age estimation, the input images are first resized to \(224\times224\). The feature maps are extracted from the original embedding of the Swin Transformer used as a backbone, and from the first three stages. The sizes of the feature maps are  \( 128 \times 56 \times 56 \), \( 256 \times 28 \times 28 \), \( 512 \times 14 \times 14 \), and \( 1024 \times 7 \times 7 \), respectively. These feature maps are forwarded to ARCM (see Section \ref{sec:ARCM}) for the enhancement of feature representation. After the feature refinement and selection of important features using attention mechanism by ARCM, feature fusion (see Section \ref{sec:FF}) is applied to concatenate all these features and produce a final feature vector of size \( 1 \times 1024 \). The feature vector is then processed through the MTAH (see Section\ref{sec:MTAH}), which produces two outputs, including both direct regression output and also a probability distribution for the age estimation. 

The model is trained using AdamW optimizer with a learning rate of \( 2 e-5 \) and weight decay of \( 5 e-3 \) on a batch size of 16 images. To enhance training stability and convergence, we apply the ReduceLROnPlateau scheduler with a factor of 0.5 and patience of 5, which reduces the learning rate when the validation MAE plateaus. A weighted combination of Huber loss with \(\delta = 1\) with KL divergence is used for the calculation of loss. To ensure robust model training, we implement K-Fold cross-validation with K = 5, saving five models. To get the final result, we report the average of outputs from theses models as the final prediction. MAE is used for the evaluation of our proposed approach for age estimation.

\section{Results and Discussion}
\label{sec:Results_Discussion}

\begin{table}[]
\centering
\scriptsize
\caption{PSNR and SSIM values calculated between the original image and the two reconstructed ones using Deep-Fillv2, after the occlusion of the eyes and the occlusion of the mouth, for each dataset. For PSNR, higher is better. For SSIM, a perfect match has a value of $1$.}
\label{tab:occlusion_metrics}
\begin{tabular}{|c|cc|cc|cc|}
\hline
\multirow{2}{*}{\textbf{Metrics}} & \multicolumn{2}{c|}{\textbf{FG-NET}}                                                                                                                      & \multicolumn{2}{c|}{\textbf{UTKFace}}                                                                                                                     & \multicolumn{2}{c|}{\textbf{MORPH}}                                                                                                                       \\ \cline{2-7} 
                                            & \multicolumn{1}{c|}{\textbf{\begin{tabular}[c]{@{}c@{}}Eyes \\ Reconstructed\end{tabular}}} & \textbf{\begin{tabular}[c]{@{}c@{}}Mouth\\ Reconstructed\end{tabular}} & \multicolumn{1}{c|}{\textbf{\begin{tabular}[c]{@{}c@{}}Eyes \\ Reconstructed\end{tabular}}} & \textbf{\begin{tabular}[c]{@{}c@{}}Mouth\\ Reconstructed\end{tabular}} & \multicolumn{1}{c|}{\textbf{\begin{tabular}[c]{@{}c@{}}Eyes \\ Reconstructed\end{tabular}}} & \textbf{\begin{tabular}[c]{@{}c@{}}Mouth\\ Reconstructed\end{tabular}} \\ \hline
PSNR                                        & \multicolumn{1}{c|}{41.26}                                                            & 37.51                                                             & \multicolumn{1}{c|}{39.20}                                                            & 34.89                                                             & \multicolumn{1}{c|}{39.26}                                                            & 35.42                                                             \\ \hline
SSIM                                        & \multicolumn{1}{c|}{0.98}                                                             & 0.96                                                              & \multicolumn{1}{c|}{0.96}                                                             & 0.91                                                              & \multicolumn{1}{c|}{0.96}                                                             & 0.89                                                              \\ \hline
\end{tabular}
\end{table}

\subsection{Occlusion Removal}
\label{sec:Occlusion_Removal_Result}

In this Section, we discuss the results obtained by occlusion removal process (see Section \ref{sec:Occlusion_Removal_Methodology})  by computing PSNR and SSIM values between original and reconstructed images. The PSNR, focused on measuring the pixel-level accuracy, and the SSIM, more oriented to grading the perceptual similarity, are utilized to evaluate the similarity between the reconstructed and the original images. Table \ref{tab:occlusion_metrics} shows the SSIM and PSNR values computed between the original and reconstructed images. The best results for both metrics, and also for eye and mouth occlusion, are achieved in the FG-NET dataset, with $41.26$ PSNR and $0.98$ SSIM. Considering that PSNR values between $30$ and $40$dB are regarded as good, produced by a high quality reconstruction, we observe that Deep-Fillv2 worked well in all datasets and cases, and even worked very well in FG-NET with occluded eyes, with a result higher than 40 dB, which is close to an almost lossless situation.

\begin{figure}
\centerline{\includegraphics[width=0.95\linewidth]{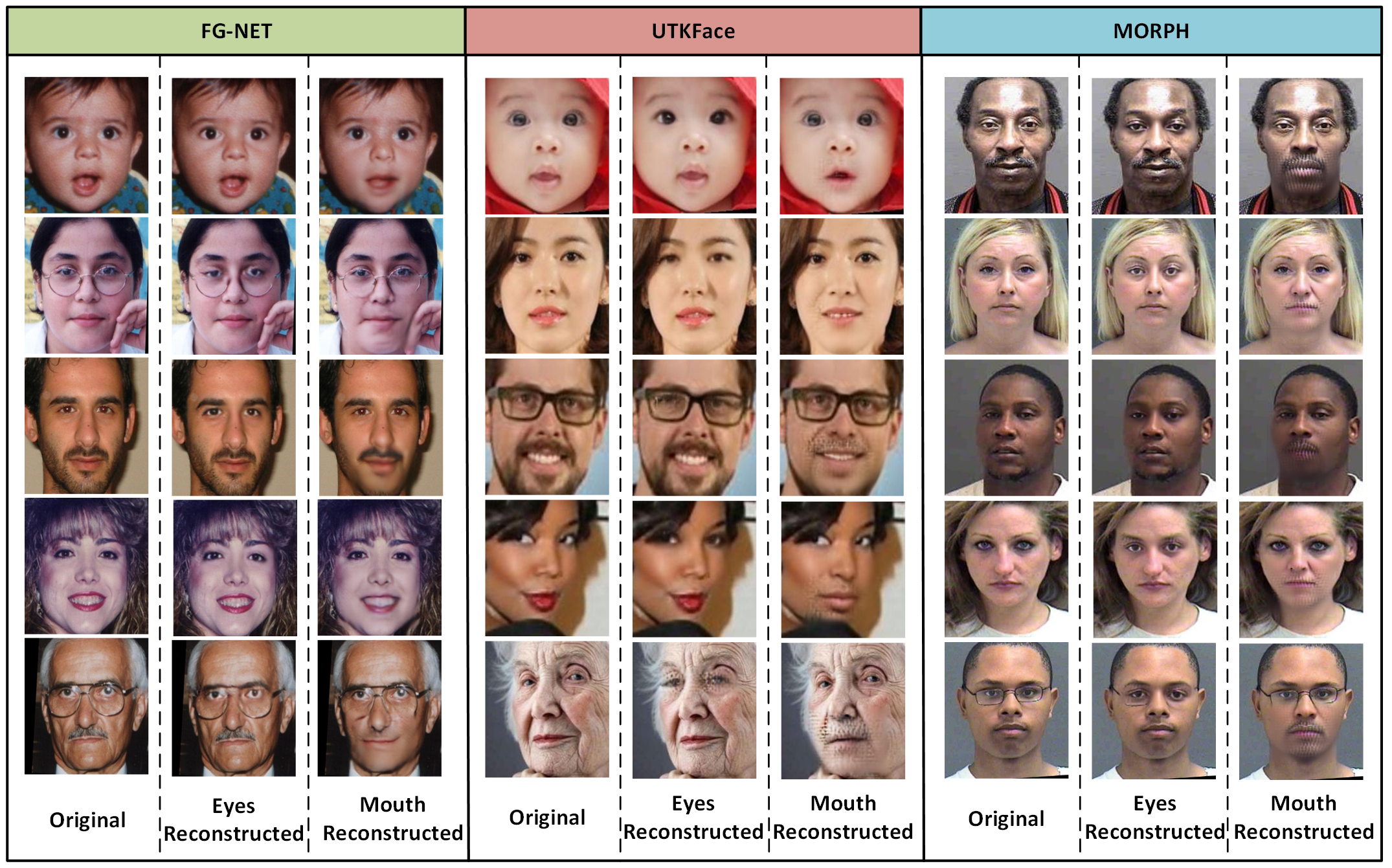}} 
\caption{Original and their corresponding reconstructed image samples obtained after removal of eye and mouth occlusion. Column 1, 2, and 3 represents image samples from FG-NET, UTKFace, and MORPH dataset respectively.}
\label{fig. OriginalvsReconstructed}
\end{figure}

In the case of SSIM, where values close to but below $0.9$ are considered good, indicating only minor distortions, this is clearly the case for the reconstructed images with the mouth occluded from MORPH, $0.89$, and UTKFace, $0.91$. In addition, it is remarkable that in all cases where the eyes are occluded, and also in FG-NET with mouth occluded, the SSIM is higher than $0.95$, which is considered excellent and almost indistinguishable from the original. Therefore, based on the calculated metrics, we can observe that both of them agree, and that the reconstruction process is good in general and even very good in FG-NET,  but let's take a look at the generated images, some of which are shown in Fig. \ref{fig. OriginalvsReconstructed}.

Among the reconstructed images, it can be observed that the eye-reconstructed images have a better aspect than the mouth-reconstructed images across all three datasets and appear almost similar to the original ones, confirming the results obtained by the metrics \ref{tab:occlusion_metrics}. This makes sense because the occluded areas introduced to cover the eyes are smaller than those of the mouth. In addition, the eye region contains more information, which helps the generative model to learn and produce richer representations that are more similar to the original ones.

In all cases, the generative model cannot guess certain aspects, such as the color of the eyes, the direction of the gaze, or even the specific shape of the eyes. In the case of the mouth, the model cannot predict the exact thickness of the lips, the opening of the mouth, or whether the person is smiling.
As the results are worse and very close in UTKFace and MORPH, it can be observed that the reasons could be the variety of poses, illumination, expressions, and races in these two datasets compared to FG-NET, which was expected and not at all surprising.
In any case, as can be seen in the images, these differences do not affect the apparent age of the individuals.

\subsection{Age Estimation}
\label{sec:Age_Estimation_Result}

\begin{table}[]
\centering
\scriptsize
\caption{Comparison of our proposal with the  state of the art methods for occluded facial age estimation as reported in the corresponding papers. Eyes and Mouth represent the eye-occluded and mouth-occluded images whereas Eyes (Reconstructed) and Mouth (Reconstructed) represent the images reconstructed from occluded images after occlusion removal. The bold values correspond to the minimum MAE for each dataset and situation (i.e. eye and mouth occlusion).}
\label{tab:SOTA_metrics}
\begin{tabular}{|l|c|c|c|c|}
\hline
\multicolumn{1}{|c|}{\multirow{2}{*}{\textbf{Methods}}}                         & \multirow{2}{*}{\textbf{Occlusion Type}} & \textbf{FG-NET} & \textbf{UTKFace} & \textbf{MORPH} \\ \cline{3-5} 
\multicolumn{1}{|c|}{}                                                          &                                          & \textbf{MAE}    & \textbf{MAE}     & \textbf{MAE}   \\ \hline
\multirow{2}{*}{SSR-Net \cite{chaves2023data}}                 & Eyes                                     & 22.08           & -                & -              \\ \cline{2-5} 
                                                                                & Mouth                                    & 18.03           & -                & -              \\ \hline
LCA-GAN \cite{nam2023lca}                                      & Mouth                                    & -               & -                & 6.64           \\ \hline
EfficientNet-B3 CNN \cite{barcic2022age}                       & Mouth                                    & 12.77           & 8.70             & -              \\ \hline
VGG-face (standard T-S) \cite{georgescu2022teacher}            & Mouth                                    & -               & 5.40             & -              \\ \hline
SSR-Net Old \cite{chaves2020improving}                         & Eyes                                     & -               & -                & 4.19           \\ \hline
\multirow{4}{*}{Baseline: Swin Transformer \cite{liu2021swin}} & Eyes                                     & 3.51            & 4.66             & 2.48           \\ \cline{2-5} 
                                                                                & Mouth                                    & 3.49            & 4.98             & 2.82           \\ \cline{2-5} 
                                                                                & Eyes (Reconstructed)                     & \textbf{3.78}   & \textbf{4.58}    & \textbf{2.61}  \\ \cline{2-5} 
                                                                                & Mouth (Reconstructed)                    & \textbf{3.43}   & \textbf{4.92}    & \textbf{2.75}  \\ \hline
\multirow{4}{*}{Ours: Swin Base + ARCM + MTAH}                                  & Eyes                                     & 3.18            & 4.64             & 2.47           \\ \cline{2-5} 
                                                                                & Mouth                                    & 3.06            & 4.95             & 2.75           \\ \cline{2-5} 
                                                                                & Eyes (Reconstructed)                     & \textbf{3.00}   & \textbf{4.54}    & \textbf{2.53}  \\ \cline{2-5} 
                                                                                & Mouth (Reconstructed)                    & \textbf{2.98}   & \textbf{4.90}    & \textbf{2.73}  \\ \hline
\end{tabular}
\end{table}

To verify the effectiveness of our proposed approach for facial age estimation (see Section \ref{sec:Age_Estimation_Methodology}), we compare it with state-of-art-methods. Table \ref{tab:SOTA_metrics} shows the comparison of our proposal with the other state of the art approaches for occluded facial age estimation. We used a Swin Transformer for age estimation as our baseline, with reconstructed and occluded images of both eyes and mouth.

We improved the baseline Swin Transformer by adding an Attentive Residual Convolution Module (ARCM), feature fusion, and a Multi-Task Age Head (MTAH). Unlike most existing methods, which only consider mouth occlusion, we have experimented with occluding both the mouth and the eyes. 
Therefore, our approach has been evaluated on the eye and mouth occluded versions of the datasets mentioned in Section \ref{sec:Datasets} both directly with the occluded images and also applying the occlusion removal explained in Section \ref{sec:Occlusion_Removal_Experimentation}. Our proposed approach outperforms all existing State-of-the-art approaches in facial age estimation with any type of occlusion across all three datasets FG-NET, UTKFace and MORPH. For all of them, our approach performed better with reconstructed images than with occluded images. In the case of the FG-NET dataset, the best results are obtained with reconstructed mouth images with an MAE of $2.98$, whereas for the UTKFace and MORPH datasets, the best results are obtained for reconstructed images of eyes with an MAE of $4.54$ and $2.53$, respectively. 

With respect to the FG-NET dataset, our approach achieved an improvement of $19.08$ years compared to SSR-Net \cite{chaves2023data} for eye occlusion. In the case of mouth occlusion, we have achieved a reduction in MAE by $15.05$ years from SSR-Net \cite{chaves2023data} and $9.79$ years from EfficientNet-B3 CNN \cite{barcic2022age}. In the case of the UTKFace dataset, no existing state-of-the-art approach has been evaluated for eye occlusion, resulting our MAE as the best so far. However, for the mouth occlusion, we achieved the best results by improving the MAE with a difference of $4.16$ years from EfficientNet-B3 CNN \cite{barcic2022age} and $0.86$ years from VGG-face (standard T-S) \cite{georgescu2022teacher}. For the MORPH dataset, our approach outperformed existing state-of-the-art methods for both eye and mouth occlusion, reducing the MAE by $1.66$ years compared to SSR-Net Old \cite{chaves2020improving} for eye occlusion and by $3.91$ years compared to LCA-GAN \cite{nam2023lca} for mouth occlusion.

\begin{table}[]
\centering
\scriptsize
\caption{Comparison of our approach for occluded facial age estimation with state-of-the-art methods for non-occluded facial age estimation, based on results reported in the corresponding papers for the FG-NET dataset. 
The $\uparrow$ expresses that our approach has obtained a higher MAE on the reconstructed images obtained after occlusion removal, compared to the SOTA method on the non-occluded images.}
\label{tab:SOTA_OC_NOC_FGNET_metrics}
\begin{tabular}{|lc|c|c|c|c|}
\hline
\multicolumn{2}{|c|}{\textbf{\begin{tabular}[c]{@{}c@{}}Methods\\ Non-Occluded\\ (MAE)\end{tabular}}} & \textbf{\begin{tabular}[c]{@{}c@{}}Ours \\ Eyes \\ Reconstructed\\ (MAE)\end{tabular}} & \textbf{\begin{tabular}[c]{@{}c@{}}Ours \\ Mouth \\ Reconstructed\\ (MAE)\end{tabular}} & \textbf{\begin{tabular}[c]{@{}c@{}}Difference \\ Eyes \\ Reconstructed\\ (MAE)\end{tabular}} & \textbf{\begin{tabular}[c]{@{}c@{}}Difference \\ Mouth \\ Reconstructed \\ (MAE)\end{tabular}} \\ \hline
\multicolumn{1}{|l|}{Ranking-CNN \cite{chen2017deep}}   & 5.79                       & \multirow{9}{*}{3.00}                                                                  & \multirow{9}{*}{2.98}                                                                   & \textbf{2.79}                                                                                & \textbf{2.81}                                                                                  \\ \cline{1-2} \cline{5-6} 
\multicolumn{1}{|l|}{DEX \cite{rothe2018deep}}          & 4.63                       &                                                                                        &                                                                                         & \textbf{1.63}                                                                                & \textbf{3.00}                                                                                  \\ \cline{1-2} \cline{5-6} 
\multicolumn{1}{|l|}{DRF \cite{shen2019deep}}           & 3.47                       &                                                                                        &                                                                                         & \textbf{0.47}                                                                                & \textbf{0.49}                                                                                  \\ \cline{1-2} \cline{5-6} 
\multicolumn{1}{|l|}{DAA \cite{chen2023daa}}            & 2.19                       &                                                                                        &                                                                                         & 0.81 $\uparrow$                                                                              & 0.79 $\uparrow$                                                                                \\ \cline{1-2} \cline{5-6} 
\multicolumn{1}{|l|}{MCGRL \cite{shou2025masked}}       & 2.10                       &                                                                                        &                                                                                         & 0.90 $\uparrow$                                                                              & 0.88 $\uparrow$                                                                                \\ \cline{1-2} \cline{5-6} 
\multicolumn{1}{|l|}{MGP-Net \cite{zang2025multi}}      & \multicolumn{1}{l|}{3.18}  &                                                                                        &                                                                                         & \textbf{0.18}                                                                                & \textbf{0.20}                                                                                  \\ \cline{1-2} \cline{5-6} 
\multicolumn{1}{|l|}{ADPF \cite{wang2022improving}}     & \multicolumn{1}{l|}{2.86}  &                                                                                        &                                                                                         & 0.14 $\uparrow$                                                                              & 0.12 $\uparrow$                                                                                \\ \cline{1-2} \cline{5-6} 
\multicolumn{1}{|l|}{AMRL \cite{zhao2022adaptive}}      & \multicolumn{1}{l|}{3.61}  &                                                                                        &                                                                                         & \textbf{0.61}                                                                                & \textbf{0.63}                                                                                  \\ \cline{1-2} \cline{5-6} 
\multicolumn{1}{|l|}{MWR \cite{shin2022moving}}         & \multicolumn{1}{l|}{2.23}  &                                                                                        &                                                                                         & 0.77 $\uparrow$                                                                              & 0.75 $\uparrow$                                                                                \\ \hline
\end{tabular}
\end{table}

Table \ref{tab:SOTA_OC_NOC_FGNET_metrics} illustrates the comparison of our proposed approach for occluded facial age estimation (see Section \ref{sec:Age_Estimation_Methodology}) against state-of-the-art (SOTA) methods for non-occluded facial age estimation on the FG-NET dataset. Despite its small size, FG-NET poses significant challenges due to variations in pose, lighting, and facial expressions, making it ideal for evaluating model generalization under constrained conditions. For eye occlusion, our proposed approach achieves an MAE of $3.00$ years, surpassing several SOTA approaches by notable margins. Specifically, it outperforms Ranking-CNN \citep{chen2017deep}, DEX \citep{rothe2018deep}, DRF \citep{shen2019deep}, MGP-Net \citep{zang2025multi}, and AMRL \citep{zhao2022adaptive} by $2.79$, $1.63$, $0.47$, $0.18$, and $0.61$ years, respectively. For mouth occlusion, our approach achieves an MAE of $2.98$ years, outperforming the same above-mentioned methods by $2.81$, $3.00$, $0.49$, $0.20$, and $0.63$ years, respectively. These improvements demonstrate our approach’s ability to effectively handle multiple types of facial occlusions, extracting reliable age-related features from the reconstructed facial areas obtained after occlusion removal (see Section \ref{sec:Occlusion_Removal_Methodology}). However, compared to other SOTA methods, our approach exhibits a slightly higher MAE, with differences of less than 1 year for both eye and mouth occlusions. These minor differences, under 1 year, highlight the robustness of our approach, especially since it tackles the more complex task of occluded age estimation, unlike SOTA methods that rely on ideal, non-occluded conditions.

\begin{table}[]
\centering
\scriptsize
\caption{Comparison of our approach for occluded facial age estimation with state-of-the-art methods for non-occluded facial age estimation, based on results reported in the corresponding papers for the UTKFace dataset. The $\uparrow$ expresses that our approach has obtained a higher MAE on the reconstructed images obtained after occlusion removal, compared to the SOTA method on the non-occluded images.}
\label{tab:SOTA_OC_NOC_UTK_New_metrics}
\begin{tabular}{|lc|c|c|c|c|}
\hline
\multicolumn{2}{|c|}{\textbf{\begin{tabular}[c]{@{}c@{}}Methods\\ Non-Occluded\\ (MAE)\end{tabular}}}   & \textbf{\begin{tabular}[c]{@{}c@{}}Ours \\ Eyes \\ Reconstructed\\ (MAE)\end{tabular}} & \textbf{\begin{tabular}[c]{@{}c@{}}Ours \\ Mouth \\ Reconstructed\\ (MAE)\end{tabular}} & \textbf{\begin{tabular}[c]{@{}c@{}}Difference \\ Eyes \\ Reconstructed\\ (MAE)\end{tabular}} & \textbf{\begin{tabular}[c]{@{}c@{}}Difference \\ Mouth \\ Reconstructed\\ (MAE)\end{tabular}} \\ \hline
\multicolumn{1}{|l|}{DEX \cite{rothe2018deep}}             & 4.31                      & \multirow{7}{*}{4.54}                                                                  & \multirow{7}{*}{4.90}                                                                   & 0.23 $\uparrow$                                                                              & 0.59 $\uparrow$                                                                               \\ \cline{1-2} \cline{5-6} 
\multicolumn{1}{|l|}{RCOR \cite{cao2020rank}}              & 5.39                      &                                                                                        &                                                                                         & \textbf{0.85}                                                                                & \textbf{0.49}                                                                                 \\ \cline{1-2} \cline{5-6} 
\multicolumn{1}{|l|}{SDTL \cite{al2020age}}                & 4.86                      &                                                                                        &                                                                                         & \textbf{0.32}                                                                                & 0.04 $\uparrow$                                                                               \\ \cline{1-2} \cline{5-6} 
\multicolumn{1}{|l|}{SAM \cite{li2020age}}                 & 4.77                      &                                                                                        &                                                                                         & \textbf{0.23}                                                                                & 0.13 $\uparrow$                                                                               \\ \cline{1-2} \cline{5-6} 
\multicolumn{1}{|l|}{MWR \cite{shin2022moving}}            & \multicolumn{1}{l|}{4.37} &                                                                                        &                                                                                         & 0.17 $\uparrow$                                                                              & 0.53 $\uparrow$                                                                               \\ \cline{1-2} \cline{5-6} 
\multicolumn{1}{|l|}{MiVOLO \cite{kuprashevich2023mivolo}} & \multicolumn{1}{l|}{4.23} &                                                                                        &                                                                                         & 0.31 $\uparrow$                                                                              & 0.67 $\uparrow$                                                                               \\ \cline{1-2} \cline{5-6} 
\multicolumn{1}{|l|}{GroupFace \cite{zhang2024groupface}}  & \multicolumn{1}{l|}{4.32} &                                                                                        &                                                                                         & 0.22 $\uparrow$                                                                              & 0.58 $\uparrow$                                                                               \\ \hline
\end{tabular}
\end{table}

Table \ref{tab:SOTA_OC_NOC_UTK_New_metrics} presents the comparison of our occluded facial age estimation approach against SOTA methods applied to non-occluded images on the UTKFace dataset. For eye occlusion, our method achieves an MAE of $4.54$ years, outperforming several SOTA methods by achieving a reduction in MAE. It surpasses RCOR \citep{cao2020rank}, SDTL \citep{al2020age}, and SAM \citep{li2020age}, by $0.85$, $0.32$, and $0.23$ years, respectively. For mouth occlusion, our approach records an MAE of $4.90$ years, achieving a reduction in MAE of $0.49$ years against RCOR \citep{cao2020rank}. These improvements highlight our approach’s capability to handle occlusions effectively, even in the complex, uncontrolled environment of UTKFace. Compared to the remaining SOTA methods, our approach shows slightly higher MAE, with differences of less than $0.4$ years for eye occlusion and less than $0.7$ years for mouth occlusion. These small differences, all under $0.7$ years, are notable, as our approach addresses the added complexity of occlusion to reflect real-world scenarios where the full face is not visible for age estimation, unlike SOTA methods, which perform only under controlled settings (i.e. non-occluded images). These results highlight the generalization and effectiveness of our approach, achieving competitive performance under occlusion in a diverse and noisy dataset, and demonstrates that our method is well suited for both controlled and uncontrolled wild scenarios.

\begin{table}[]
\centering
\scriptsize
\caption{Comparison of our approach for occluded facial age estimation with state-of-the-art methods for non-occluded facial age estimation, based on results reported in the corresponding papers for the MORPH dataset. The $\uparrow$ expresses that our approach has obtained a higher MAE on the reconstructed images obtained after occlusion removal, compared to the SOTA method on the non-occluded images.}
\label{tab:SOTA_OC_NOC_MORPH_metrics}
\begin{tabular}{|ll|c|c|c|c|}
\hline
\multicolumn{2}{|c|}{\textbf{\begin{tabular}[c]{@{}c@{}}Methods\\ Non-Occluded\\ (MAE)\end{tabular}}}  & \textbf{\begin{tabular}[c]{@{}c@{}}Ours \\ Eyes \\ Reconstructed\\ (MAE)\end{tabular}} & \textbf{\begin{tabular}[c]{@{}c@{}}Ours \\ Mouth \\ Reconstructed\\ (MAE)\end{tabular}} & \textbf{\begin{tabular}[c]{@{}c@{}}Difference \\ Eyes \\ Reconstructed\\ (MAE)\end{tabular}} & \textbf{\begin{tabular}[c]{@{}c@{}}Difference \\ Mouth \\ Reconstructed \\ (MAE)\end{tabular}} \\ \hline
\multicolumn{1}{|l|}{Ranking-CNN \cite{chen2017deep}}     & \multicolumn{1}{c|}{2.96} & \multirow{10}{*}{2.53}                                                                 & \multirow{10}{*}{2.73}                                                                  & \textbf{0.43}                                                                                & \textbf{0.23}                                                                                  \\ \cline{1-2} \cline{5-6} 
\multicolumn{1}{|l|}{DEX \cite{shen2019deep}}             & \multicolumn{1}{c|}{2.68} &                                                                                        &                                                                                         & \textbf{0.15}                                                                                & 0.05 $\uparrow$                                                                                \\ \cline{1-2} \cline{5-6} 
\multicolumn{1}{|l|}{CNN+ELM \cite{duan2017ensemble}}     & \multicolumn{1}{c|}{4.03} &                                                                                        &                                                                                         & \textbf{1.53}                                                                                & \textbf{1.30}                                                                                  \\ \cline{1-2} \cline{5-6} 
\multicolumn{1}{|l|}{RAGN \cite{duan2017ensemble}}        & \multicolumn{1}{c|}{2.61} &                                                                                        &                                                                                         & \textbf{0.08}                                                                                & 0.12 $\uparrow$                                                                                \\ \cline{1-2} \cline{5-6} 
\multicolumn{1}{|l|}{MWR \cite{shin2022moving}}           & 2.00                      &                                                                                        &                                                                                         & 0.53 $\uparrow$                                                                              & 0.73 $\uparrow$                                                                                \\ \cline{1-2} \cline{5-6} 
\multicolumn{1}{|l|}{MCGRL \cite{shou2025masked}}         & 1.89                      &                                                                                        &                                                                                         & 0.64 $\uparrow$                                                                              & 0.84 $\uparrow$                                                                                \\ \cline{1-2} \cline{5-6} 
\multicolumn{1}{|l|}{MGP-Net \cite{zang2025multi}}        & 2.37                      &                                                                                        &                                                                                         & 0.16 $\uparrow$                                                                              & 0.36 $\uparrow$                                                                                \\ \cline{1-2} \cline{5-6} 
\multicolumn{1}{|l|}{DAA \cite{chen2023daa}}              & 2.06                      &                                                                                        &                                                                                         & 0.47 $\uparrow$                                                                              & 0.67 $\uparrow$                                                                                \\ \cline{1-2} \cline{5-6} 
\multicolumn{1}{|l|}{ADPF \cite{wang2022improving}}       & 2.54                      &                                                                                        &                                                                                         & \textbf{0.01}                                                                                & 0.19 $\uparrow$                                                                                \\ \cline{1-2} \cline{5-6} 
\multicolumn{1}{|l|}{GroupFace \cite{zhang2024groupface}} & 1.86                      &                                                                                        &                                                                                         & 0.67 $\uparrow$                                                                              & 0.87 $\uparrow$                                                                                \\ \hline
\end{tabular}
\end{table}

Table \ref{tab:SOTA_OC_NOC_MORPH_metrics} illustrates the comparison of our occluded facial age estimation approach against SOTA methods applied to non-occluded images on the MORPH dataset. The large size of the data set and its demographic diversity make it a robust platform to evaluate the scalability and ability of a model to generalize across diverse populations. For eye occlusion, our method achieves an MAE of $2.53$ years, surpassing several SOTA methods by achieving reductions in MAE. It outperforms Ranking-CNN \citep{chen2017deep}, DEX \citep{shen2019deep}, CNN+ELM \citep{duan2017ensemble}, RAGN \citep{duan2017ensemble}, and ADPF \citep{wang2022improving}, by $0.43$, $0.15$, $1.53$, $0.08$, and $0.01$ years, respectively. For mouth occlusion, our method records an MAE of $2.73$ years, achieving reductions in MAE of $0.23$ years against Ranking-CNN \citep{chen2017deep} and $1.30$ years against CNN+ELM \citep{duan2017ensemble}. These improvements demonstrate our approach’s ability to effectively mitigate the impact of occlusions on a large and diverse dataset. Compared to the remaining SOTA methods, our approach shows slightly higher MAE, with differences of less than $0.7$ years for eye occlusion and less than $1$ year for mouth occlusion. These slight differences highlight our approach's competitive performance despite the occlusion challenge. The ability to achieve such close results on MORPH, a dataset with extensive demographic and age coverage, underscores the scalability and robustness of our method.

\subsection{Scalability Analysis}
\label{sec:Scalability_Analysis}
We have performed a scalability analysis to demonstrate the effectiveness of our ARCM and MTAH modules when added to Swin transformers of increasing complexity on the FG-NET, UTKFace, and MORPH datasets.
Table \ref{tab:Ablation Study} illustrates the comparison of our occluded facial age estimation approach using Swin-Tiny, Swin-Small, and Swin-Large as backbones, evaluated on the FG-NET, UTKFace, and MORPH datasets under eye and and mouth occlusion scenarios. As expected, our approach, using Swin-Large as the backbone, outperforms the other variants by achieving the lowest MAE values on all three datasets for both eye and mouth occlusions. For eye occlusion, MAEs of 2.85, 4.48, and 2.42 years are achieved on the FG-NET, UTKFace, and MORPH datasets, respectively. For mouth occlusion, MAEs of 2.95, 4.82, and 2.70 years are achieved on the same datasets, respectively. In contrast, all other Swin transformer variants exhibit higher MAE values for both eye and mouth occlusions, with Swin-Tiny performing least effectively, recording MAEs of 3.35 and 3.16 years (FG-NET), 4.79 and 5.11 years (UTKFace), and 2.56 and 2.84 years (MORPH) for eye and mouth occlusions, respectively.

\begin{table}[]
\centering
\scriptsize
\caption{Evaluation of our proposed approach with different backbone architectures of Swin Transformer for FG-NET, UTKFace, and MORPH datasets.}
\label{tab:Ablation Study}
\begin{tabular}{|c|l|c|c|c|}
\hline
\textbf{Models}                           & \multicolumn{1}{c|}{\textbf{Occlusion Type}} & \textbf{\begin{tabular}[c]{@{}c@{}}FG-NET\\ (MAE)\end{tabular}} & \textbf{\begin{tabular}[c]{@{}c@{}}UTKFace \\ (MAE)\end{tabular}} & \textbf{\begin{tabular}[c]{@{}c@{}}MORPH \\ (MAE)\end{tabular}} \\ \hline
\multirow{2}{*}{Swin-Tiny + ARCM + MTAH}  & Eyes Reconstructed                           & 3.35                                                            & 4.79                                                              & 2.56                                                            \\ \cline{2-5} 
                                          & Mouth Reconstructed                          & 3.16                                                            & 5.11                                                              & 2.84                                                            \\ \hline
\multirow{2}{*}{Swin-Small + ARCM + MTAH} & Eyes Reconstructed                           & 3.23                                                            & 4.74                                                              & 2.49                                                            \\ \cline{2-5} 
                                          & Mouth Reconstructed                          & 3.13                                                            & 5.09                                                              & 2.81                                                            \\ \hline
\multirow{2}{*}{Swin-Base + ARCM + MTAH}  & Eyes Reconstructed                           & 3.00                                                            & 4.54                                                              & 2.53                                                            \\ \cline{2-5} 
                                          & Mouth Reconstructed                          & 2.98                                                            & 4.90                                                              & 2.73                                                            \\ \hline
\multirow{2}{*}{Swin-Large + ARCM + MTAH} & Eyes Reconstructed                           & \textbf{2.85}                                                   & \textbf{4.48}                                                     & \textbf{2.42}                                                   \\ \cline{2-5} 
                                          & Mouth Reconstructed                          & \textbf{2.95}                                                   & \textbf{4.82}                                                     & \textbf{2.70}                                                   \\ \hline
\end{tabular}
\end{table}

The superior performance of Swin-Large's 24 layers architecture is due to its larger initial embedding dimension of 192, and final feature vector size of 1536, compared to Swin-Base’s 24 layers, initial embedding of 128, and feature vector of 1024, Swin-Small’s 24 layers, initial embedding of 96, and feature vector of 768, and Swin-Tiny’s 12 layers, initial embedding of 96, and feature vector of 768. With 197 million parameters, exceeding Swin-Base’s 88 million, Swin-Small’s 50 million, and Swin-Tiny’s 28 million, Swin-Large variant effectively captures complex hierarchical features critical for age estimation under occlusion. Although Swin-Tiny and Swin-Small offer computational efficiency, their smaller feature vectors and fewer parameters limit accuracy, highlighting a trade-off between efficiency and performance.

\begin{figure}
\centerline{\includegraphics[height=  5 cm, width=11 cm]{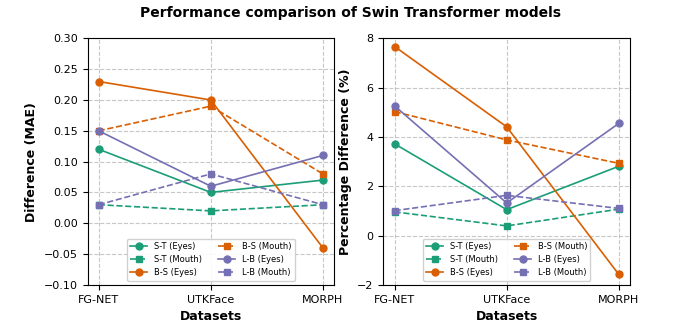}}
\caption{Comparison of performance between Swin Transformer variants.}
\label{fig:Swin_Comparison}
\end{figure}

Fig. \ref{fig:Swin_Comparison} illustrates the performance comparison between Swin Transformer variants for both eye and mouth occlusions across FG-NET, UTKFace, and MORPH datasets. L-B, B-S, and S-T represent comparisons of absolute and percentage differences between the Large and Base, Base and Small, and Small and Tiny models, respectively. For eye occlusions, Swin-L outperforms Swin-B, Swin-S, and Swin-T across all datasets by capturing complex features. On FG-NET, a small dataset with significant variability in aging dynamics and image quality, Swin-L achieves a 0.25 absolute and 8.00\% gain, surpassing Swin-B (0.23, 7.67\%), Swin-S (0.18, 6.50\%), and Swin-T (0.15, 5.80\%) due to its robust modeling of variable aging cues, while Swin-B struggles with noise, Swin-S lacks generalization, and Swin-T underfits small, complex data. On UTKFace, which contains diverse images that vary in age, gender, race, and lighting, Swin-L records a gain of 0.22 absolute and 4.80\%, outperforming Swin-B (0.20, 4.41\%), Swin-S (0.16, 3.90\%), and Swin-T (0.13, 3.50\%) by leveraging diverse patterns. Swin-B misses fine-grained details, Swin-S struggles with subtle variations, and Swin-T lacks the capacity to represent the dataset’s diversity. For MORPH, which contains a large number of frontal and well-lit images in a controlled environment, Swin-L gains 0.12 absolute and 4.70\%, outperforming Swin-B (0.11, 4.55\%), Swin-S (0.09, 4.20\%), and Swin-T (0.08, 4.00\%) by extracting subtle age cues, while smaller models miss these due to limited capacity, although they perform better when age-related features are more pronounced. For mouth occlusions, Swin-L achieves 5.20\%, 4.00\%, and 3.00\% gains on FG-NET, UTKFace, and MORPH, exceeding Swin-B (5.03\%, 3.88\%, 2.93\%), Swin-S (4.80\%, 3.60\%, 2.70\%), and Swin-T (4.50\%, 3.30\%, 2.50\%), respectively due to effective feature learning despite fewer age cues.

\subsection{Qualitative Analysis}

\begin{figure}[]
\centerline{\includegraphics[width=0.70\linewidth]{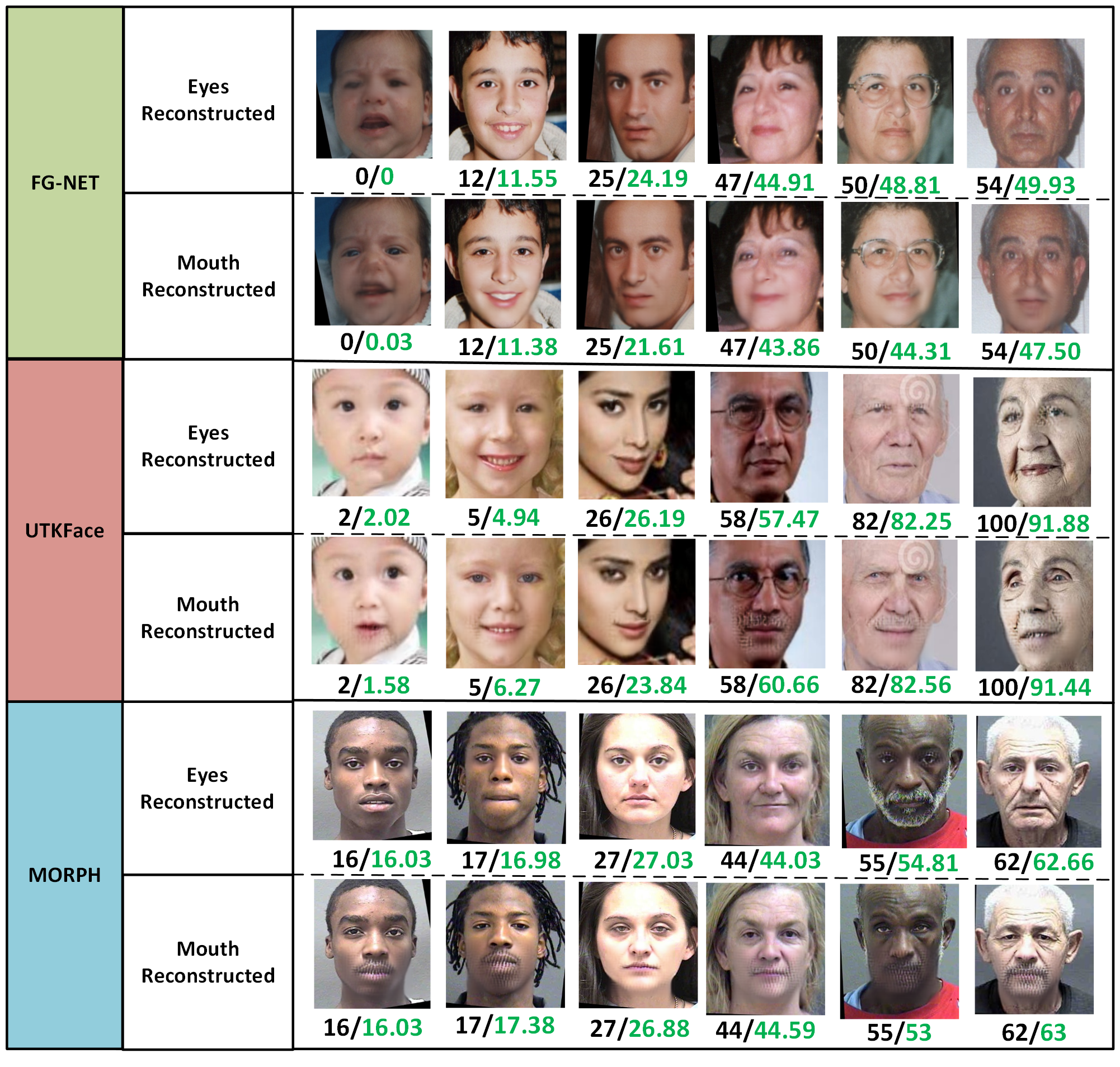}} 
\caption{Samples of reconstructed eye and mouth images after occlusion removal from FG-NET, UTKFace, and MORPH datasets, with actual and estimated ages. Ages in black indicate actual ages, while ages in green show estimated ages from our proposed approach.}
\label{fig:Qualitative_Analysis}
\end{figure}


We randomly selected six samples of reconstructed eye and mouth images from each of the FG-NET, UTKFace, and MORPH datasets to visualize the prediction results of our proposed approach against ground truth values. As Fig. \ref{fig:Qualitative_Analysis} shows, our proposed approach performed better for the reconstructed eye images compared to the reconstructed mouth images across all three datasets. This is because we achieved better reconstruction results for eye-occluded images compared to mouth-occluded images. It can be observed that the reconstructed eye images closely resemble normal images, whereas blurriness can be seen in the reconstructed mouth images. This results in a degrading effect on the teeth, lips, chin, and surrounding areas, causing larger age estimation errors in the reconstructed mouth images.

Despite the challenges in mouth reconstruction, our approach achieved better age estimation results for most age groups across all three datasets, except in a few cases where the difference between the actual and estimated age exceeded 5 years. The distribution of image samples for FG-NET, UTKFace, and MORPH datasets is shown in Fig. \ref{fig:Combined_Distribution}. The FG-NET dataset primarily consists of samples from ages 0–40 years. Our approach performed better for age groups under 50 years but struggled with ages above 50 years due to the imbalance of the dataset. A similar pattern is observed in the UTKFace dataset, which contains most samples from ages 1–90 years. Our approach performed well across all age groups in the UTKFace dataset, except for the oldest groups (i.e., ages above 91 years). The MORPH dataset has a better sample distribution for ages 16–59 years but fewer samples for ages above 59 years. Our approach performed better for the age range of 16–69 years, with the difference between the actual and estimated age not exceeding 2 years.

\section{Conclusions}

In this work, we presented a novel framework for the occluded facial age estimation to enhance its applicability in uncontrolled wild scenarios. We applied SN-Patch GAN for removing the eye and mouth occlusions from the occluded images. We proposed ARCM coupled with a Swin Transformer for the feature extraction and enhancement of feature representation. We also introduced an MTAH module for effective age estimation by using the combination of both regression and distribution learning. Our approach outperforms the existing state-of-the-art methods for occluded facial age estimation on three benchmark datasets (FG-NET, UTKFace, and MORPH). We also compared our approach for occluded facial age estimation with SOTA methods developed for non-occluded facial age estimation to demonstrate the effectiveness of our approach in bridging the gap between occluded and non-occluded age estimation. %
Our approach reduces the performance gap between occluded and non-occluded facial age estimation to less than 1 year on the FG-NET and MORPH datasets, and to less than 0.7 years on the UTKFace dataset, making it suitable for uncontrolled real-world scenarios.
\subsection{Limitations}

Following are some of the limitations of our work:

\begin{itemize}
    \item \textit{Severe occlusion:} Although our approach outperforms SOTA methods in occluded facial age estimation, there is still room for improvement. It handles partial occlusions well but may struggle with severe cases where the eyes, nose, and mouth are all occluded. However, such scenarios are rare.

    \item \textit{Variable Masking:} Our approach currently uses rectangular masking to create masks for occlusion removal. In cases of variable occlusion, which is not fixed and can occur on any part of the face in any shape other than rectangular, our approach has not been assessed in those scenarios.

\end{itemize}

\subsection{Future Work}

To address the limitations mentioned above, we suggest the following ideas for future work:

\begin{itemize}
    \item \textit{Occlusion removal:} We suggest designing an effective GAN-based or difussion-based model that can reconstruct images from limited visible information.

    \item \textit{Adaptive Masking:} Our suggestion is to develop dynamic masking strategies, such as irregular or contour-based masks, using segmentation models to mimic real-world occlusion shapes (e.g., shadows or objects).
    
\end{itemize} 

\section*{Data Availability}

Data will be made available upon request.

\section*{Acknowledgments}

This work has been funded by the Recovery, Transformation, and Resilience Plan, financed by the European Union (Next Generation), thanks to the LUCIA project (Fight against Cybercrime by applying Artificial Intelligence) granted by INCIBE to the University of Le\'on.


\bibliographystyle{IEEEtranN}
\bibliography{bibliography}

\end{document}